\documentclass[letterpaper]{article}
\usepackage{times}
\usepackage{helvet}
\usepackage{courier}
\usepackage{url}
\usepackage{graphicx}
\usepackage{subfigure}
\usepackage{algorithm}
\usepackage{booktabs}
\usepackage{algorithmic}
\usepackage[]{amsmath}
\usepackage{amsfonts}
\usepackage{mathpazo}

\usepackage[]{hyphenat}
\usepackage[]{aaai17}
\usepackage{todonotes}
\usepackage[makeroom]{cancel}

\DeclareMathOperator*{\E}{\mathbb E}
\DeclareMathOperator*{\Var}{Var}

\title{Mean Actor-Critic}

\author{
Cameron Allen$^{1}$\thanks{These authors contributed equally. Please send correspondence to Cameron Allen \texttt{<csal@brown.edu>} and Kavosh Asadi \texttt{<kavosh@brown.edu>}.} \
Kavosh Asadi$^{1*}$ \
Melrose Roderick$^{1}$ \
Abdel-rahman Mohamed$^{2}\thanks{This work was completed while at Microsoft Research.}$ \\
{\bf \Large George Konidaris$^{1}$ \ Michael Littman$^{1}$ }\\
\\
\begin{tabular}{ccc}
Brown University$^{1}$ & \quad\quad\quad & Amazon$^{2}$ \\
Providence, RI & \quad\quad\quad & Seattle, WA\\
\end{tabular}
}

\begin{document}

\maketitle
\begin{abstract}
We propose a new algorithm, Mean Actor-Critic (MAC), for discrete-action continuous-state reinforcement learning. MAC is a policy gradient algorithm that uses the agent's explicit representation of all action values to estimate the gradient of the policy, rather than using only the actions that were actually executed. We prove that this approach reduces variance in the policy gradient estimate relative to traditional actor-critic approaches. We show empirical results on two control domains and six Atari games, where MAC is competitive with state-of-the-art policy search methods.
\end{abstract}

\section{Introduction}
In reinforcement learning (RL), two important classes of algorithms are value-function-based methods and policy search methods. Value-function-based methods maintain an estimate of the value of performing each action in each state, and choose the actions associated with the most value in their current state \cite{sutton1998reinforcement}. By contrast, policy search algorithms maintain an explicit policy, and agents draw actions directly from that policy to interact with their environment \cite{sutton2000policy}. A subset of policy search algorithms, policy gradient methods, represent the policy using a differentiable parameterized function approximator (for example, a neural network) and use stochastic gradient ascent to update its parameters to achieve more reward.

To facilitate gradient ascent, the agent interacts with its environment according to the current policy and keeps track of the outcomes of its actions. From these (potentially noisy) sampled outcomes, the agent estimates the gradient of the objective function. A critical question here is how to compute an accurate gradient using these samples, which may be costly to acquire, while using as few sample interactions as possible.

Actor-critic algorithms compute the policy gradient using a learned value function to estimate expected future reward \cite{sutton2000policy,konda2000actor}. Since the expected reward is a function of the environment's dynamics, which the agent does not know, it is typically estimated by executing the policy in the environment. Existing algorithms compute the policy gradient using the value of states the agent visits, and critically, these methods take into account only the actions the agent actually executes during environmental interaction.

We propose a new policy gradient algorithm, Mean Actor-Critic (or MAC), for the discrete-action continuous-state case. MAC uses the agent's policy distribution to \textit{average the value function over all actions}, rather than using the action-values of only the sampled actions. We prove that, under modest assumptions, this approach reduces variance in the policy gradient estimates relative to traditional actor-critic approaches. We implement MAC using deep neural networks, and we show empirical results on two control domains and six Atari games, where MAC is competitive with state-of-the-art policy search methods.

We note that the core idea behind MAC has also been independently and concurrently explored by \citeauthor{ciosek2017expected} \shortcite{ciosek2017expected}. However, their results mainly focus on continuous action spaces and are more theoretical. We introduce a simpler proof of variance reduction that makes fewer assumptions, and we also show that the algorithm works well in discrete-action domains.



\section{Background}
In RL, we train an agent to select actions in its environment so that it maximizes some notion of long-term reward. We formalize the problem as a Markov decision process (MDP) \cite{puterman1990markov}, which we specify by the tuple $\langle \mathcal{S},s_0,\mathcal{A,R,T},\gamma\rangle$, where $\mathcal{S}$ is a set of states, $s_0\in\mathcal{S}$ is a fixed initial state, $\mathcal{A}$ is a set of discrete actions, the functions $\mathcal{R}: \mathcal{S \times A}\rightarrow \mathbb{R}$ and
$\mathcal{T: S \times A \times S\rightarrow}[0,1]$ respectively describe the reward and transition dynamics of the environment, and $\gamma \in [0,1)$ is a discount factor representing the relative importance of immediate versus long-term rewards.

More concretely, we denote the expected reward for performing action $a \in \mathcal{A}$ in state $s \in \mathcal{S}$ as:
$$\mathcal{R}(s,a)= \E \big[r_{t+1}  \big|s_t=s,a_t=a\big]\ ,$$
and we denote the probability that performing action $a$ in state $s$ results in state $s' \in \mathcal{S}$ as:
$$\mathcal{T}(s,a,s')= \textrm{Pr}(s_{t+1}=s'\big|s_t=s,a_t=a)\ .$$

In the context of policy search methods, the agent maintains an explicit policy $\pi(a|s;{\theta})$ denoting the probability of taking action $a$ in state $s$ under the policy $\pi$ parameterized by $\theta$. Note that for each state, the policy outputs a probability distribution over the discrete set of actions: $\pi:\mathcal{S}\rightarrow \mathcal{P}(\mathcal{A})$. At each timestep $t$, the agent takes an action $a_t$ drawn from its policy $\pi(\cdot|s_t;\theta)$, then the environment provides a reward signal $r_t$ and transitions to the next state $s_{t+1}$.

The agent's goal at every timestep is to maximize the sum of discounted future rewards, or simply \textit{return}, which we define as:
\begin{equation}
G_t=\sum_{k=1}^{\infty} \gamma^{k-1}r_{t+k} \ . \nonumber
\end{equation}

In a slight abuse of notation, we will also denote the total return for a trajectory $\tau$ as $G(\tau)$, which is equal to $G_0$ for that same trajectory.

The agent's policy induces a value function over the state space. The expression for return allows us to define both a state value function, $V^\pi(s)$, and a state-action value function, $Q^\pi(s,a)$. Here, $V^\pi(s)$ represents the expected return starting from state $s$, and following the policy $\pi$ thereafter, and $Q^\pi(s,a)$ represents the expected return starting from $s$, executing action $a$, and then following the policy $\pi$ thereafter:
$$V^\pi (s) := \E_{\pi}\big[G_t\big|s_t=s\big],$$
$$Q^\pi (s,a) := \E_{\pi}\big[G_t\big|s_t=s, a_t=a\big].$$
Note that:
$$V^\pi(s) = \sum_{a \in \mathcal{A}}\ [\pi(a|s;\theta)Q^\pi(s,a)].$$

The agent's goal is to find a policy that maximizes the return for every timestep, so we define an objective function $J$ that allows us to score an arbitrary policy parameter $\theta$:
\begin{eqnarray*}
J(\theta)&=&\E_{\tau\sim Pr(\tau|\theta)}[G(\tau)]=\sum_{\tau} Pr(\tau|\theta)G(\tau)\ ,
\label{spg_obj}
\end{eqnarray*}
where $\tau$ denotes a trajectory. Note that the probability of a specific trajectory depends on policy parameters as well as the dynamics of the environment. Our goal is to be able to compute the gradient of $J$ with respect to the policy parameters $\theta$:
\begin{eqnarray}
\nabla_{\theta}J(\theta)&=&\sum_{\tau} \nabla_{\theta}Pr(\tau|\theta)G(\tau) \nonumber \\
&=&\sum_{\tau} Pr(\tau|\theta)\frac{\nabla_{\theta}Pr(\tau|\theta)}{Pr(\tau|\theta)}G(\tau)\nonumber\\
&=&\sum_{\tau} Pr(\tau|\theta)\nabla_{\theta}\log Pr(\tau|\theta)G(\tau)\nonumber\\
&=&\E_{s\sim d^{\pi},\ a\sim \pi}[\nabla_{\theta}\log \pi(a|s;\theta)G_0]\nonumber\\
&=&\E_{s\sim d^{\pi},\ a\sim \pi}[\nabla_{\theta}\log \pi(a|s;\theta)G_t]\nonumber\\
&=&\E_{s\sim d^{\pi},\ a\sim \pi}[\nabla_{\theta}\log \pi(a|s;\theta)Q^{\pi}(s,a)]
\label{eq:spg_grad_default}
\end{eqnarray}
where $d^\pi(s)=\sum_{t=0}^{\infty}\gamma^t Pr(s_t = s | s_0, \pi)$ is the discounted state distribution. In the second and third lines we rewrite the gradient term using a score function. In the fourth line, we convert the summation to an expectation, and use the $G_0$ notation in place of $G(\tau)$. Next, we make use of the fact that $\E[G_0] = \E[G_t]$, given by \citeauthor{williams1992simple} \shortcite{williams1992simple}. Intuitively this makes sense, since the policy for a given state should depend only on the rewards achieved after that state. Finally, we invoke the definition that $Q^{\pi}(s,a) = \E[G_t]$.

A nice property of expectation~(\ref{eq:spg_grad_default}) is that, given access to $Q^{\pi}$, the expectation can be estimated through implementing policy $\pi$ in the environment. Alternatively, we can estimate $Q^{\pi}$ using the return $G_t$, which is an unbiased (and usually a high variance) sample of $Q^{\pi}$. This is essentially the idea behind the REINFORCE algorithm \cite{williams1992simple}, which uses the following gradient estimator:
\begin{equation}
\nabla_{\theta} J(\theta) \approx \frac{1}{T} \sum_{t=1}^{T} G_t\nabla_{\theta}\log \pi(a_t|s_t;\theta) .
\label{eq:REINFORCE_grad_estimation}
\end{equation}
Alternatively, we can estimate $Q^\pi$ using some sort of function approximation: $\widehat Q(s,a;\omega) \approx Q^{\pi}(s,a)$, which results in variants of actor-critic algorithms. Perhaps the simplest actor-critic algorithm approximates~(\ref{eq:spg_grad_default}) as follows:
\begin{equation}
\nabla_{\theta} J(\theta) \approx \frac{1}{T} \sum_{t=1}^{T} \widehat Q(s_t,a_t;w)\nabla_{\theta}\log \pi(a_t|s_t;\theta).
\label{eq:ac_grad_estimation}
\end{equation}
Note that value function approximation can, in general, bias the gradient estimation \cite{baxter2001infinite}.

One way of reducing variance in both REINFORCE and actor-critic algorithms is to use an additive control variate as a baseline \cite{williams1992simple,sutton2000policy,greensmith2004variance}. The baseline function is typically a function that is fixed over actions, and so subtracting it from either the sampled returns or the estimated Q-values does not bias the gradient estimation. We refer to techniques that use such a baseline as \textit{advantage} variations of the basic algorithms, since they approximate the advantage $A(s,a)$ of choosing action $a$ over some baseline representing ``typical" performance for the policy in state $s$ \cite{baird1994reinforcement}. The update performed by advantage REINFORCE is:
\begin{equation}
\theta \leftarrow \theta +\alpha \sum_{t=1}^{T}(G_t-b) \nabla_{\theta} \log \pi(a_t|s_t;\theta) \ , \nonumber
\end{equation}
where $b$ is a scalar baseline measuring the performance of the policy, such as a running average of the observed return over the past few episodes of interaction.

Advantage actor-critic uses an approximation of the expected value of each state $s_t$ as its baseline: $\widehat V(s_t) := \sum_a{\pi(a|s_t;\theta)\widehat Q(s_t,a;\omega)}$, which leads to the following update rule:
\begin{equation}
\theta \leftarrow \theta +\alpha \sum_{t=1}^{T}\big(\widehat Q(s_t,a_t;\omega)-\widehat V(s_t)\big)\nabla_{\theta} \log \pi(a_t|s_t;\theta) \ . \nonumber
\end{equation}

Another way of estimating the advantage function is to use the TD-error signal $\delta = r_t + \gamma V(s') - V(s)$. This approach is convenient, because it only requires estimating one set of parameters, namely for $V$. However, because the TD-error is a sample of the advantage function $A(s,a) = Q^{\pi}(s,a) - V^{\pi}(s)$, this approach has higher variance (due to the environmental dynamics) than methods that explicitly compute $Q(s,a) - V(s)$. Moreover, given $Q$ and $\pi$, $V$ can easily be computed as $V=\sum_{a}\pi(a|s) Q(s,a)$, so in practice, it is still only necessary to estimate one set of parameters (for $Q$).

\section{Mean Actor-Critic}
An overwhelming majority of recent actor-critic papers have computed the policy gradient using an estimate similar to Equation~(\ref{eq:ac_grad_estimation}) \cite{degris2012off,mnih2016asynchronous,wang2016sample}. This estimate samples both states and actions from trajectories executed according to the current policy in order to compute the gradient of the objective function with respect to the policy weights.

Instead of using only the sampled actions, Mean Actor-Critic (MAC) explicitly computes the probability-weighted average over all Q-values, for each state sampled from the trajectories. In doing so, MAC is able to produce an estimate of the policy gradient where the variance due to action sampling is reduced to zero. This is exactly the difference between computing the sample mean (whose variance is inversely proportional to the number of samples), and calculating the mean directly (which is simply a scalar with no variance).

MAC is based on the observation that expectation~(\ref{eq:spg_grad_default}), which we repeat here, can be rewritten in the following way:
\begin{align}
\nabla_{\theta}J(\theta)&=\E_{s\sim d^{\pi},\ a\sim \pi}[\nabla_{\theta}\log \pi(a|s;\theta)Q^{\pi}(s,a)]\nonumber\\
&=\E_{s \sim d^{\pi}}\Big[\sum_{a \in \mathcal{A}}\pi(a|s;\theta)\nabla_{\theta}\log \pi(a|s;\theta)Q^{\pi}(s,a)\Big]\nonumber\\
&=\E_{s\sim d^{\pi}}\Big[\sum_{a \in \mathcal{A}}\nabla_{\theta}\pi(a|s;\theta)Q^{\pi}(s,a)\Big].
\label{eq:mac_expectation}
\end{align}

We can estimate~(\ref{eq:mac_expectation}) by sampling states from a trajectory and using function approximation:
 \begin{equation*}
 	\nabla_{\theta}J(\theta)\approx \frac{1}{T} \sum_{t=0}^{T-1}\sum_{a \in \mathcal{A}}\nabla_{\theta}\pi(a|s_t;\theta)\widehat Q(s_t,a;\omega)\ .
 \end{equation*}

In our implementation, the inner summation is computed by combining two neural networks that represent the policy and state-action value function. The value function can be learned using a variety of methods, such as temporal-difference learning or Monte Carlo sampling. After performing a few updates to the value function, we update the parameters $\theta$ of the policy with the following update rule:
 \begin{equation}
\theta \leftarrow \theta +\alpha \sum_{t=0}^{T-1}\sum_{a \in \mathcal{A}}\nabla_{\theta}\pi(a|s_t;\theta)\widehat Q(s_t,a;\omega).
\label{eq:mac_update_rule}
\end{equation}

To improve stability, repeated updates to the value and policy networks are interleaved, as in Generalized Policy Iteration \cite{sutton1998reinforcement}.

In traditional actor-critic approaches, which we refer to as \textit{sampled-action} actor-critic, the only actions involved in the computation of the policy gradient estimate are those that were actually executed in the environment. In MAC, computing the policy gradient estimate will frequently involve actions that were not actually executed in the environment. This results in a trade-off between bias and variance. In domains where we can expect accurate Q-value predictions from our function approximator, despite not actually executing all of the relevant state-action pairs, MAC results in lower variance gradient updates and increased sample-efficiency. In domains where this assumption is not valid, MAC may perform worse than sampled-action actor-critic due to increased bias.

\begin{figure}[t]
\centering
\parbox{3.5cm}{
{%
\setlength{\fboxsep}{0pt}%
\setlength{\fboxrule}{0.5pt}%
\fbox{\includegraphics[width=3cm]{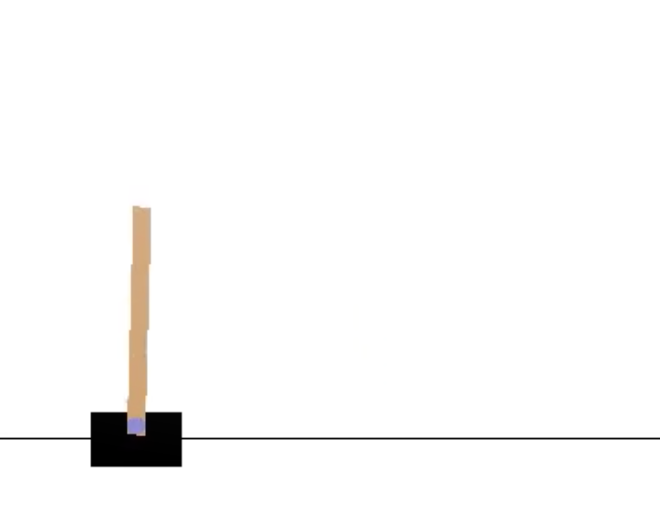}}
}%
\label{fig:cart_screenshot}}
\qquad
\begin{minipage}{3.5cm}
{%
\setlength{\fboxsep}{0pt}%
\setlength{\fboxrule}{0.5pt}%
\fbox{\includegraphics[width=3cm]{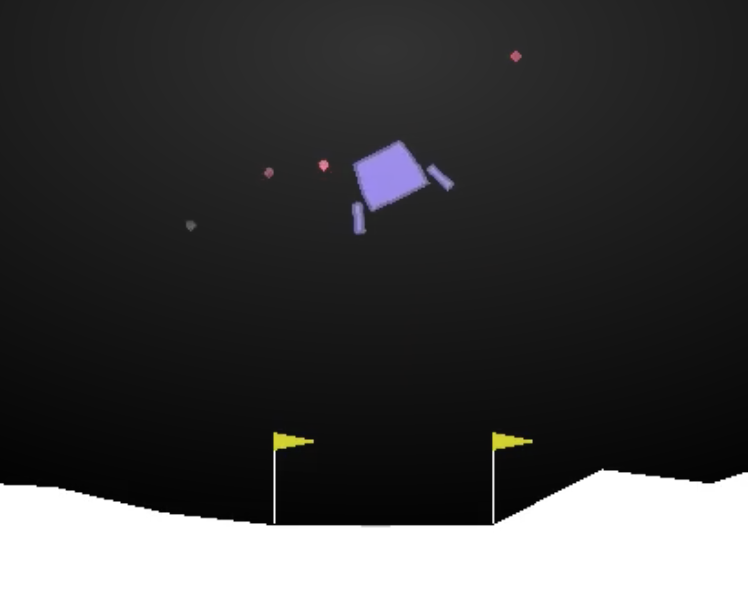}}
}%
\label{fig:lunar_screenshot}
\end{minipage}
\caption{Screenshots of the classic control domains Cart Pole (left) and Lunar Lander (right)}
\label{fig:control_screenshots}
\end{figure}

In some ways, MAC is similar to Expected Sarsa \cite{van2009theoretical}. Expected Sarsa considers all next-actions $a_{t+1}$, then computes the expected TD-error, $\E[\delta] = r_t + \gamma\E[Q(s_{t+1}, a_{t+1})] - Q(s_t, a_t)$, and uses the resulting error signal to update the $Q$ function. By contrast, MAC considers all current-actions $a_t$, and uses the corresponding $Q(s_t,a_t)$ values to update the policy directly.

It is natural to consider whether MAC could be improved by subtracting an action-independent baseline, as in sampled-action actor-critic and REINFORCE:
\begin{equation*}
\nabla_{\theta}J(\theta)=\E_{s\sim d^{\pi}}\Big[\sum_{a \in \mathcal{A}}\nabla_{\theta}\pi(a|s;\theta)\Big(Q^{\pi}(s,a) - V^\pi(s)\Big)\Big]. \nonumber
\end{equation*}
However, we can simplify the expectation as follows:
\begin{align*}
\nabla_{\theta}J(\theta)=\E_{s\sim d^{\pi}}\Big[\sum_{a \in \mathcal{A}}\nabla_{\theta}&\pi(a|s;\theta)Q^{\pi}(s,a)\nonumber\\
   & - V^\pi(s) \nabla_{\theta}\sum_{a \in \mathcal{A}}\pi(a|s;\theta)\Big]. \nonumber
\end{align*}
In doing so, we see that both $V^\pi(s)$ and the gradient operator can be moved outside of the summation, leaving just the sum of the action probabilities, which is always 1, and hence the gradient of the baseline term is always zero. This is true regardless of the choice of baseline, since the baseline cannot be a function of the actions or else it will bias the expectation. Thus, we see that subtracting a baseline is unnecessary in MAC, since it has no effect on the policy gradient estimate.

\section{Analysis of Bias and Variance}
In this section we prove that MAC does not increase variance over sampled-action actor-critic (AC), and also, that given a fixed $\widehat{Q}$, both algorithms have the same bias. We start with the bias result.

\subsection{Theorem 1}
If the estimated Q-values, $\widehat{Q}(s,a;\omega)$, for both MAC and AC are the same in expectation, then the bias of MAC is equal to the bias of AC.

\subsection{Proof}
See Appendix A.
\\
\\
This result makes sense because in expectation, AC will choose all of the possible actions with some probability according to the policy. MAC simply calculates this expectation over actions explicitly. We now move to the variance result.

\subsection{Theorem 2}
If the estimated Q-values, $\widehat{Q}(s,a;\omega)$, for both MAC and AC are the same in expectation, and if $\widehat{Q}(s,a;\omega)$ is independent of $\widehat{Q}(s',a';\omega)$ for $(s,a) \neq (s',a')$, then $\text{Var}[\text{MAC}]$ $\leq$ $\text{Var}[\text{AC}]$. For deterministic policies, there is equality, and for stochastic policies the inequality is strict.
\subsection{Proof}
See Appendix B.
\\
\\
Intuitively, we can see that for cases where the policy is deterministic, MAC's formulation of the policy gradient is exactly equivalent to AC, and hence we can do no better than AC. For high-entropy policies, MAC will beat AC in terms of variance.

\section{Experiments}
This section presents an empirical evaluation of MAC across three different problem domains. We first evaluate the performance of MAC versus popular policy gradient benchmarks on two classic control problems. We then evaluate MAC on a subset of Atari 2600 games and investigate its performance compared to state-of-the-art policy search methods.

\subsection{Classic Control Experiments}
In order to determine whether MAC's lower variance policy gradient estimate translates to faster learning, we chose two classic control problems, namely Cart Pole and Lunar Lander, and compared MAC's performance against four standard sampled-action policy gradient algorithms. We used the open-source implementations of Cart Pole and Lunar Lander provided by OpenAI Gym \cite{DBLP:journals/corr/BrockmanCPSSTZ16}, in which both domains have continuous state spaces and discrete action spaces. Screenshots of the two domains are provided in Figure \ref{fig:control_screenshots}.

\begin{table}[t]
  \centering
  \begin{tabular}{lll}
    \toprule
    \textbf{Algorithm} & \textbf{Cart Pole} & \textbf{Lunar Lander} \\
    \midrule
    REINFORCE & $109.5 \pm 13.3$ & $101.1 \pm 10.5$ \\
    Adv. REINFORCE & $121.8 \pm 11.2$ & $114.7 \pm 8.1$ \\
    Actor-Critic & $138.7 \pm 13.2$ & $124.6\pm 5.1$ \\
    Adv. Actor-Critic & $157.4 \pm 6.4$ & $162.8\pm 14.9$ \\
    \textbf{MAC} & $\textbf{178.3} \pm \textbf{7.6}$ & $\textbf{163.5} \pm \textbf{12.8}$ \\
    \bottomrule\\
  \end{tabular}
  \caption{Performance summary of MAC vs. sampled-action policy gradient algorithms. Scores denote the mean performance of each algorithm over all trials and episodes.}
  \label{tbl:control_results}
\end{table}

\begin{figure*}[t]
\centering
\parbox{8cm}{
\includegraphics[width=8cm]{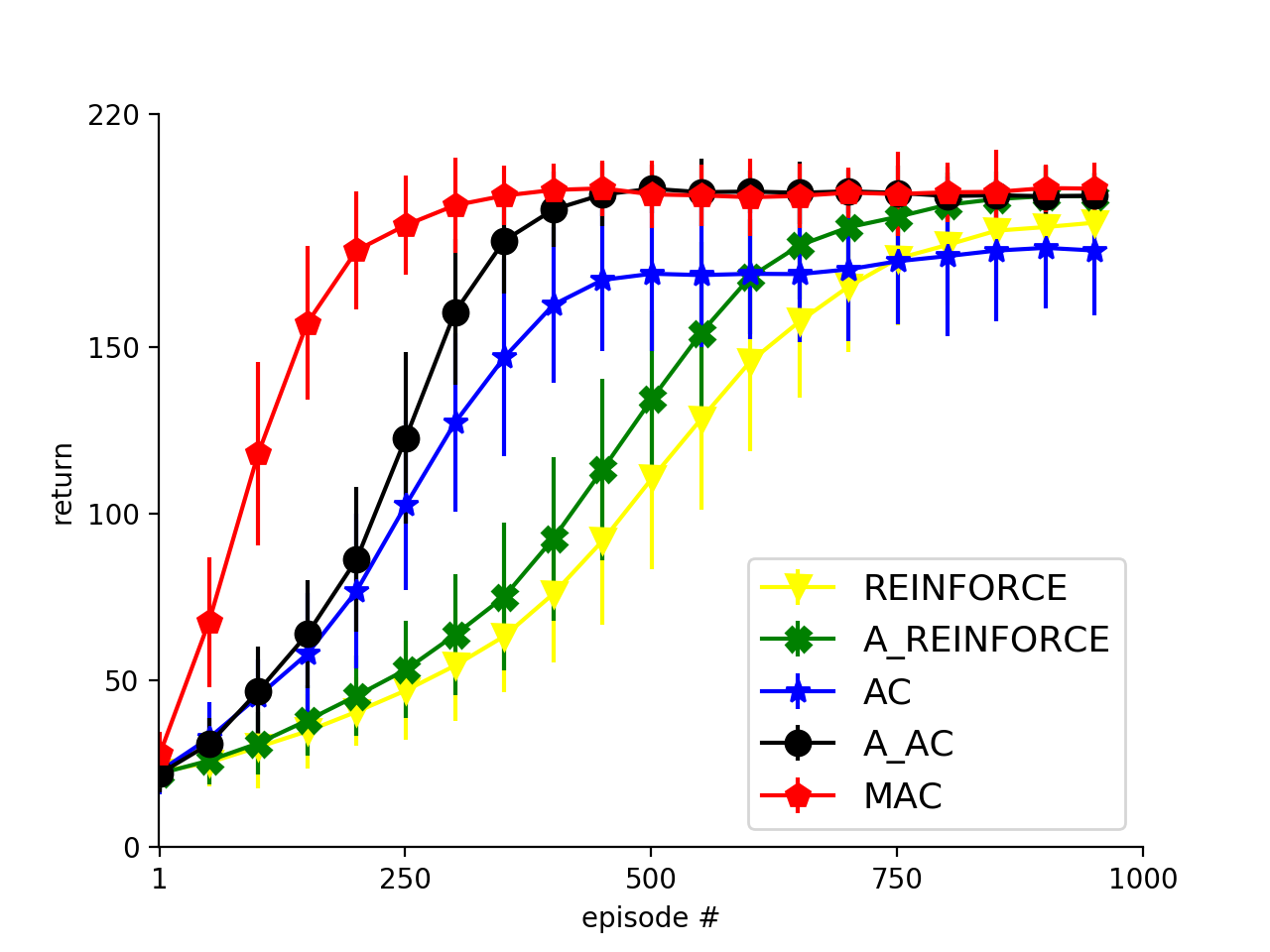}
\label{fig:cart_results}}
\qquad
\begin{minipage}{8cm}
\includegraphics[width=8cm]{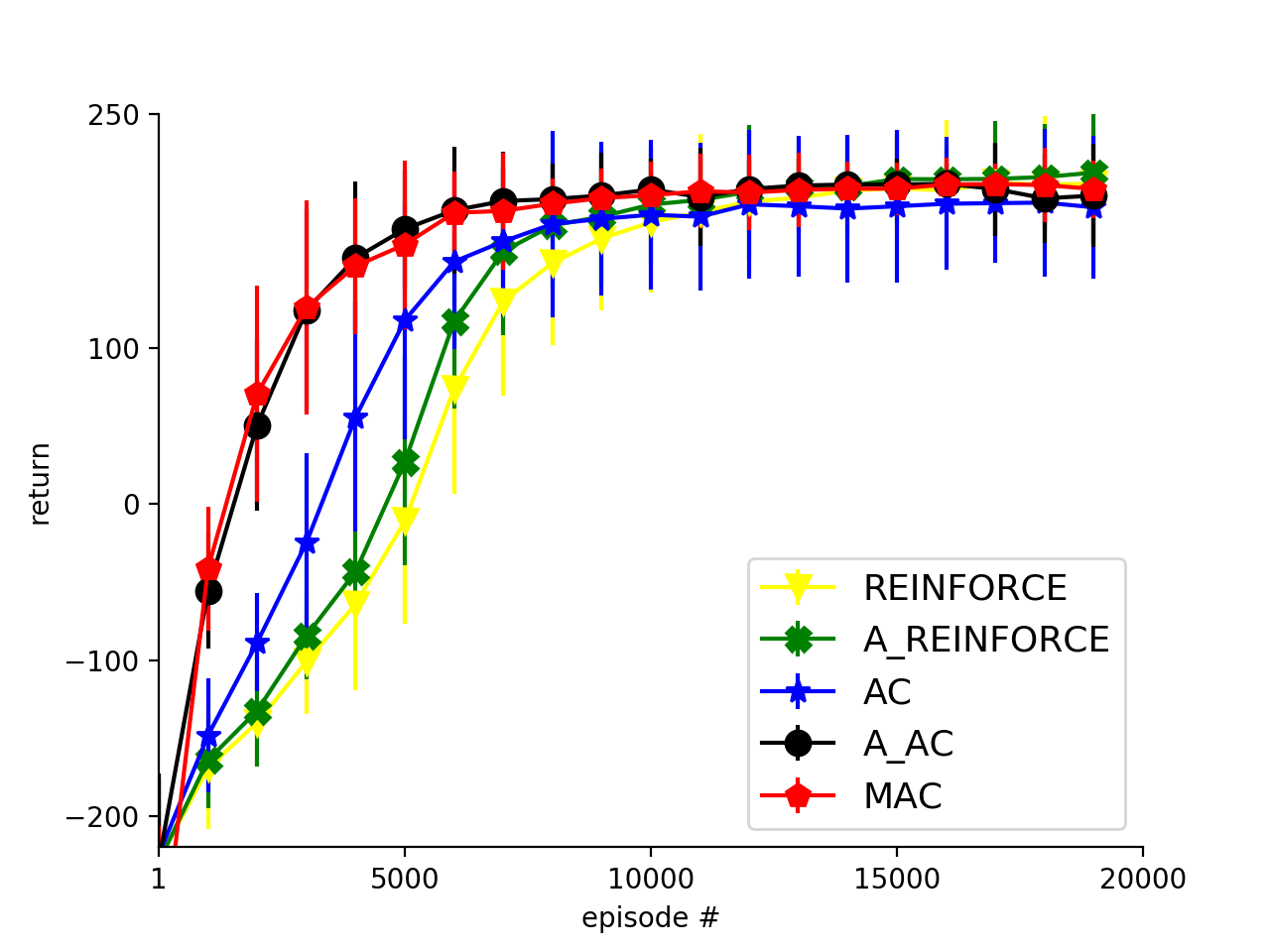}
\label{fig:lunar_results}
\end{minipage}
\caption{Performance comparison for CartPole (left) and Lunar Lander (right) of MAC vs. sampled-action policy gradient algorithms. Results are averaged over 100 independent trials.}
\label{fig:control_results}
\end{figure*}

For each problem domain, we implemented MAC using two independent neural networks, representing the policy and Q function. We then performed a hyperparameter search to determine the best network architectures, optimization method, and learning rates. Specifically, the hyperparameter search considered: 0, 1, 2, or 3 hidden layers; 50, 75, 100, or 300 neurons per layer; ReLU, Leaky ReLU (with leak factor 0.3), or tanh activation; SGD, RMSProp, Adam, or Adadelta as the optimization method; and a learning rate chosen from 0.0001, 0.00025, 0.0005, 0.001, 0.005, 0.01, or 0.05. To find the best setting, we ran 10 independent trials for each combination of hyperparameters and chose the setting with the best asymptotic performance over the 10 trials. We terminated each episode after 200 and 1000 timesteps (in Cart Pole and Lunar Lander, respectively), regardless of the state of the agent.

We compared MAC against four standard benchmarks: REINFORCE, advantage REINFORCE, actor-critic, and advantage actor-critic. We implemented the REINFORCE benchmarks using just a single neural network to represent the policy, and we implemented the actor-critic benchmarks using two networks to represent both the policy and Q function. For each benchmark algorithm, we then performed the same hyperparameter search that we had used for MAC.

In order to keep the variance as low as possible for the advantage actor-critic benchmark, we explicitly computed the advantage function $A(s,a) = Q(s,a) - V(s)$, where $V(s) = \sum_a \pi(a|s) Q(s,a)$, rather than sampling it using the TD-error (see Section 2).

Once we had determined the best hyperparameter settings for MAC and each of the benchmark algorithms, we then ran each algorithm for 100 independent trials. Figure \ref{fig:control_results} shows learning curves for the different algorithms, and Table \ref{tbl:control_results} summarizes the results using the mean performance over trials and episodes. On Cart Pole, MAC learns substantially faster than all of the benchmarks, and on Lunar Lander, it performs competitively with the best benchmark algorithm, advantage actor-critic.

\subsection{Atari Experiments}
To test whether MAC can scale to larger problem domains, we evaluated it on several Atari 2600 games using the Arcade Learning Environment (ALE) \cite{bellemare13arcade} and compared MAC's performance against that of state-of-the-art policy search methods, namely, Trust Region Policy Optimization (TRPO) \cite{schulman2015trust}, Evolutionary Strategies (ES) \cite{salimans2017evolution}, and Advantage Actor-Critic (A2C) \cite{wu2017scalable}. Due to the computational load inherent in training deep networks to play Atari games, we limited our experiments to a subset of six Atari games: Beamrider, Breakout, Pong, Q*bert, Seaquest and Space Invaders. These six games are commonly selected for tuning hyperparameters \cite{mnih2015human,mnih2016asynchronous,wu2017scalable}, and thus provide a fair comparison against established benchmarks, despite our limited computational resources.

The MAC network architecture was derived from the OpenAI Baselines implementation of A2C \cite{wu2017scalable}. It uses three convolutional layers (size/stride/filters: 8/4/32, 4/2/64, 3/1/64), followed by a fully-connected layer (size 512), all with ReLU activation. A final fully-connected layer is split into two batches of N outputs each, where N is the number of actions. One batch uses a linear activation and corresponds to the Q-values; the other batch uses a softmax activation and corresponds to the policy. We used this architecture for both the MAC results and the A2C results. The TRPO and ES results are taken from their respective papers.

\begin{table*}[t]
\centering
\begin{tabular}{lrrrrrrrr}
    \toprule
Game           & \multicolumn{1}{c}{Random} & \multicolumn{1}{c}{TRPO} & \multicolumn{1}{c}{ES} & \multicolumn{1}{c}{A2C} & \multicolumn{1}{c}{\textbf{MAC}} \\
\cmidrule{1-6}
Beam Rider     & 363.9  & 1425.2          & 744.0         & 5846.0          & \textbf{6072.0} \\
Breakout       & 1.7    & 10.8            & 9.5           & 370.9           & \textbf{372.7} \\
Pong           & -20.7  & 20.9            & \textbf{21.0} & 18.0            & 10.6 \\
Q*bert         & 183.0  & \textbf{1973.5} & 147.5         & 1651.5          & 243.4 \\
Seaquest       & 68.4   & \textbf{1908.6} & 1390.0        & 1702.5          & 1703.4 \\
Space Invaders & 148.0  & 568.4           & 678.5         & \textbf{1201.2} & 1173.1 \\
   \bottomrule
\end{tabular}
\caption{Atari performance of MAC vs. policy search methods (random start condition). TRPO and ES results are from their respective papers \cite{schulman2015trust,salimans2017evolution}. A2C and MAC results were obtained with modified versions of the OpenAI Baselines implementation of A2C \cite{wu2017scalable}.}
\label{tbl:atari_random_start}
\end{table*}

We trained the network using a variation of the multi-part loss function used in A2C \cite{wu2017scalable}. The value loss at each timestep was equal to the mean squared error between the observed reward and the Q-value of the selected action. The policy entropy loss was simply the negative entropy of the policy at each timestep. For the A2C experiments, the policy improvement loss was the negative log probability of the selected action times its advantage value. For the MAC experiments, the policy improvement loss became the negative sum of action probabilities times their associated Q-values. The overall loss function was a linear combination of the policy improvement loss (coefficient 0.1), policy entropy loss (coefficient 0.001), and value loss (coefficient 0.5), and the network was trained using RMSProp with a learning rate of 1.5e-3. These coefficients trade off the importance of learning good Q-values, improving the policy, and preventing the policy from converging prematurely. This configuration of hyperparameters was found to perform well experimentally for both methods after a small hyperparameter search. The only difference between the A2C and MAC implementations was to replace A2C's sampled-action policy improvement loss with MAC's sum-over-actions loss; the algorithms used exactly the same architecture and hyperparameters.

For A2C and MAC, we trained a network for each game on 50 million frames of play, across 16 parallel threads, pausing every 200K frames to evaluate performance and compute learning curves. In each evaluation, we ran 16 agents in parallel, for 4500 frames (5 minutes) each, or 50 total episodes, whichever came first, and averaged the scores of the completed (or timed-out) episodes. Agents were trained and evaluated under the typical random start condition, where the game is initialized with a random number of no-op ALE actions (between 0 and 30) \cite{mnih2015human}. The A2C and MAC results in Table \ref{tbl:atari_random_start} come from the final evaluation after all 50M frames, and they are averaged across 5 trials involving separately trained networks. Learning curves for each game can be found in Figure \ref{fig:atari_learning_curves} in the Appendix. In addition to A2C, we also compared MAC against TRPO (results from a single trial) \cite{schulman2015trust}, and ES (results averaged over 30 trials) \cite{salimans2017evolution}, and found that MAC performed competitively with all three benchmark algorithms.

Note that MAC's performance on Pong and Q*bert was low relative to A2C. For Pong this was due to one of the five MAC trials obtaining a final score of -20.1 and pulling the average performance down significantly. The individual Pong scores for MAC were \{20.5, 19.7, 18.3, 14.7, -20.1\}; the scores for A2C were \{19.4, 19.4, 19.3, 16.3, 15.6\}. For Q*bert, the performance for both algorithms was much more variable. A2C scored 0.0 on 3 out of 5 trials, and MAC scored 0.0 on 2 out of 5 trials. The reason A2C's average score is so much higher than MAC's is that it had one lucky trial where it scored 7780.9 points. The individual Q*bert scores for MAC were \{557.4, 504.7, 155.1, 0.0, 0.0\}; the scores for A2C were \{7780.9, 476.6, 0.0, 0.0, 0.0\}. Additional hyperparameter tuning might lead to improved performance; however, the purpose of this Atari experiment was mainly to show that MAC is competitive with state-of-the-art policy search algorithms, and these results seem to indicate that it is.

\section{Discussion}
At its core, MAC offers a new way of computing the policy gradient that can substantially reduce variance and increase learning speed. There are a number of orthogonal improvements to policy gradient methods, such as using natural gradients \cite{kakade2002natural,peters2008natural}, off-policy learning \cite{wang2016sample,gu2016q,asadi2016sample}, second-order methods
\cite{furmston2016approximate}, and asynchronous exploration \cite{mnih2016asynchronous}. We have not investigated how MAC performs with these extensions; however, just as these improvements were added to basic actor-critic methods, they could be added to MAC as well, and we expect they would improve its performance in a similar way.

A typical use-case for actor-critic algorithms is for problem domains with continuous actions, which are awkward for
value-function-based methods \cite{sutton1998reinforcement}. One approach to dealing with continuous actions is Deterministic Policy Gradients (DPG) \cite{silver2014deterministic,lillicrap2015continuous}, which uses a deterministic policy to perform off-policy policy gradient updates. However, in settings where on-policy learning is necessary, using a deterministic policy leads to sub-optimal behavior \cite{sutton1998reinforcement}, and hence a stochastic policy is typically used instead. The recently-introduced Expected Policy Gradients (EPG) \cite{ciosek2017expected} addresses this problem by generalizing DPG for stochastic policies. However, while EPG has good experimental performance on domains with continuous action spaces, the authors do not provide experimental results for discrete domains. MAC's discrete results and EPG's continuous results are in some sense complementary.

\section{Conclusion}

\begin{figure*}[t]
\centering
\includegraphics[width=16cm]{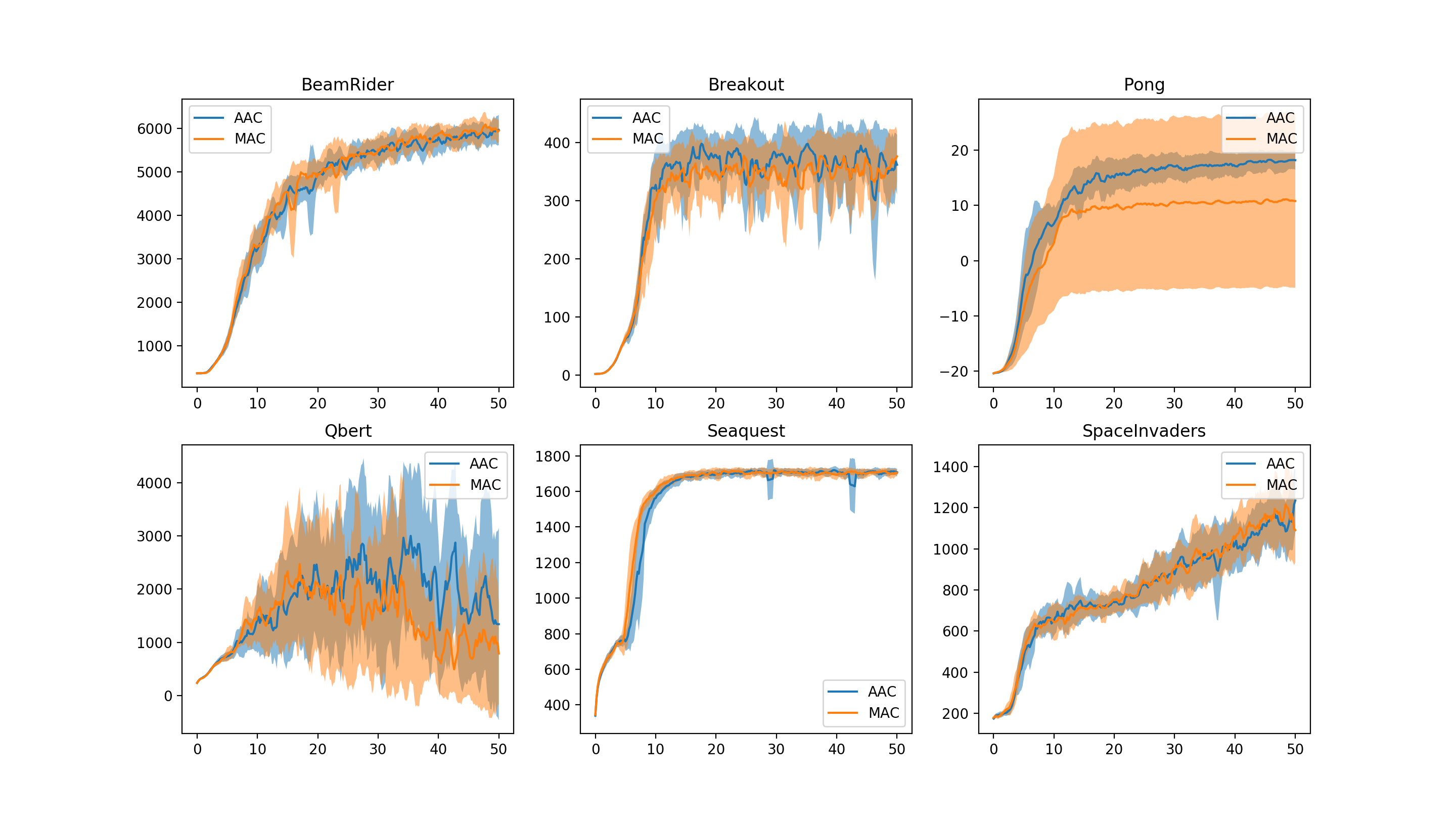}
\label{fig:atari_learning_curves}
\caption{Learning curves on six Atari games for A2C (blue) and MAC (orange). Vertical axis is score; horizontal axis is number of training frames (in millions). Results are averaged over 5 independent trials, and smoothed slightly for readability. Error bars represent standard deviation.}
\end{figure*}

The basic formulation of policy gradient estimators presented here---where the  gradient is estimated by averaging the state-action value function across actions---leads to a new family of actor-critic algorithms. This family has the advantage of not requiring an additional variance-reduction baseline, substantially reducing the design effort required to apply them. It is also a natural fit with deep neural network function approximators, resulting in a network architecture that is identical to some sampled-action actor-critic algorithms, but with less variance.

We prove that for stochastic policies, the MAC algorithm (the simplest member of the resulting family), reduces variance relative to traditional actor-critic approaches, while maintaining the same bias. Our neural network implementation of MAC either outperforms, or is competitive with, state-of-the-art policy search algorithms, and our experimental results show that MAC's lower variance lead to dramatically faster training in some cases. In future work, we aim to develop this family of algorithms further by including typical elaborations of the basic actor-critic architecture like natural or second-order gradients. Our results so far suggest that our new approach is highly promising, and that extensions to it will provide even further improvement in performance.

\bibliography{mac_refs.bib}
\bibliographystyle{aaai}

\onecolumn
\section{Appendix}
{
\setlength{\parindent}{0pt}
\setlength{\parskip}{8pt}
\raggedbottom

\subsection{A. Proof of Theorem 1}
\label{Proof:Theorem2}
Both AC and MAC are estimators of the true policy gradient (PG). Given a batch of data $D$, we can write the bias of AC and MAC as:
\begin{align}
    \text{Bias}[\text{AC}] &= \E_{\ \quad D}[AC] - \text{PG}\label{Eqn:Bias_def_AC}\\
    \text{Bias}[\text{MAC}] &= \E_{\ \quad D}[MAC] - \text{PG}\label{Eqn:Bias_def_MAC}
\end{align}
For clarity, we will rewrite the AC and MAC expectations (\ref{eq:spg_grad_default}) and (\ref{eq:mac_expectation}) to explicitly denote the way that each algorithm estimates the policy gradient, given a batch of data $D$ (with size $|D|$):
\begin{align}
\text{AC} &= \frac{1}{|D|}\sum_{(s,a)\in D} \nabla_{\theta} \log \pi(a|s;\theta) \widehat{Q}(s,a;\omega)
\label{Eqn:AC}\\
\text{MAC} &= \frac{1}{|D|}\sum_{s\in D} \sum_{a\in\mathcal{A}} \pi(a|s;\theta) \nabla_{\theta} \log \pi(a|s;\theta) \widehat{Q}(s,a;\omega)
\label{Eqn:MAC}
\end{align}
Substituting (\ref{Eqn:AC}) and (\ref{eq:spg_grad_default}) into Eqn. (\ref{Eqn:Bias_def_AC}) gives:
\begin{align}
    \text{Bias}[\text{AC}] &= \E_{\ \quad D}\Big[\frac{1}{|D|}\sum_{t = 1}^{|D|} \nabla_{\theta} \log \pi(a_t|s_t;\theta) \widehat{Q}(s_t,a_t;\omega)\Big] - \E_{s\sim d^{\pi},\ a\sim \pi}\Big[\nabla_{\theta}\log \pi(a|s;\theta)Q^{\pi}(s,a)\Big]\label{eqn:subst_ac}
\end{align}
Since $D$ is sampled from trajectories that were carried out according to the policy, we can drop the dependence on $t$ inside the expectation, and rewrite (\ref{eqn:subst_ac}) as follows:
\begin{align}
    \text{Bias}[\text{AC}] &= \frac{1}{|D|}\sum_{t = 1}^{|D|}\Big( \E_{s\sim d^{\pi},\ a\sim \pi}\Big[\nabla_{\theta} \log \pi(a|s;\theta) \widehat{Q}(s,a;\omega)\Big]\Big) - \E_{s\sim d^{\pi},\ a\sim \pi}\Big[\nabla_{\theta}\log \pi(a|s;\theta)Q^{\pi}(s,a)\Big]\\
    &= \E_{s\sim d^{\pi},\ a\sim \pi}\Big[\nabla_{\theta} \log \pi(a|s;\theta) \Big(\widehat{Q}(s,a;\omega)\ - Q^{\pi}(s,a)\Big)\Big]\\
    &= \E_{s\sim d^{\pi}}\Big[\sum_{a\in\mathcal{A}}\pi(a|s;\theta)\nabla_{\theta} \log \pi(a|s;\theta) \Big(\widehat{Q}(s,a;\omega)\ - Q^{\pi}(s,a)\Big)\Big]\label{Eqn:Bias_AC}
\end{align}
Now we turn our attention to MAC, and substitute (\ref{Eqn:MAC}) and (\ref{eq:spg_grad_default}) into Eqn. (\ref{Eqn:Bias_def_MAC}), to obtain:
\begin{align}
    \text{Bias}[\text{MAC}] &= \E_{\ \quad D}\Big[\frac{1}{|D|}\sum_{t = 1}^{|D|} \sum_{a\in\mathcal{A}} \pi(a|s_t;\theta) \nabla_{\theta} \log \pi(a|s_t;\theta) \widehat{Q}(s_t,a;\omega)\Big] - \E_{s\sim d^{\pi},\ a\sim \pi}\Big[\nabla_{\theta}\log \pi(a|s;\theta)Q^{\pi}(s,a)\Big]\label{eqn:subst_mac}\\
    &= \frac{1}{|D|}\sum_{t = 1}^{|D|}\quad \E_{s\sim d^{\pi}}\Big[\sum_{a\in\mathcal{A}} \pi(a|s;\theta) \nabla_{\theta} \log \pi(a|s;\theta) \widehat{Q}(s,a;\omega)\Big] - \E_{s\sim d^{\pi},\ a\sim \pi}\Big[\nabla_{\theta}\log \pi(a|s;\theta)Q^{\pi}(s,a)\Big]\\
    &= \E_{s\sim d^{\pi}}\Big[\sum_{a\in\mathcal{A}} \pi(a|s;\theta) \nabla_{\theta} \log \pi(a|s;\theta) \widehat{Q}(s,a;\omega)\Big] - \E_{s\sim d^{\pi}}\Big[ \sum_{a\in\mathcal{A}}\pi(a|s;\theta) \nabla_{\theta}\log \pi(a|s;\theta)Q^{\pi}(s,a)\Big]\\
    &= \E_{s\sim d^{\pi}}\Big[\sum_{a\in\mathcal{A}} \pi(a|s;\theta) \nabla_{\theta} \log \pi(a|s_t;\theta) \Big(\widehat{Q}(s,a;\omega) - Q^{\pi}(s,a)\Big)\Big]\label{Eqn:Bias_MAC}
\end{align}
Comparing (\ref{Eqn:Bias_AC}) and (\ref{Eqn:Bias_MAC}), we see that AC and MAC have the same bias.

\subsection{B. Proof of Theorem 2}
\label{Proof:Theorem3}
For any random variable $Z$, the variance $\Var[Z]$ can be written as:
$$ \Var[Z] = \E\big[Z^2\big] - \E[Z]^2 $$

If we assume the estimated Q-values for MAC and AC are the same in expectation, then the squared expectation's contribution to the variance of each algorithm will be equal. We are only interested in determining which estimator has lower variance, so we can drop the second term and simply compare $\E\big[Z^2\big]$, the second moments.

Again we will employ the explicit definitions of the AC and MAC estimators, for a data set $D$, given by (\ref{Eqn:AC}) and (\ref{Eqn:MAC}), respectively.

For ease of notation, we define the following two functions:
\begin{align}
X(s,a) &= \nabla_{\theta_i} \log \pi(a|s;\theta) \widehat{Q}(s,a;\omega) \label{Eqn:X}\\
Y(s) &= \E_{\quad\pi}\big[X(s,a)\big] = \sum_{a\in U(s)} \pi(a|s;\theta) X(s,a) \label{Eqn:H}
\end{align}

Here, $\theta_i$ represents a single parameter of the parameter vector $\theta$. We consider an arbitrary choice of $i$, so the following proof holds for all $i$.

The above expressions allow us to rewrite the AC and MAC estimators (Eqn. \ref{Eqn:AC} \& \ref{Eqn:MAC}) in terms of $X(s,a)$ and $Y(s)$:
\begin{align}
\text{AC}_i &= \frac{1}{|D|}\sum_{(s,a)\in D} X(s,a) \label{Eqn:AC_GH}\\
\text{MAC}_i &= \frac{1}{|D|}\sum_{s\in D} \sum_{a\in U(s)} Y(s) \label{Eqn:MAC_GH}
\end{align}

For convenience, we drop the $i$ subscript for the rest of this analysis.

Now we are ready to compare $\E_{s,a}[\text{AC}^2]$ vs. $\E_{s}[\text{MAC}^2]$.
\small
\begin{equation*}
\begin{split}
\E_{s,a}[\text{AC}^2] &= \E_{s,a}\Big[\Big(\frac{1}{|D|}\sum_{(s,a)\in D} X(s,a)\Big)\Big(\frac{1}{|D|}\sum_{(s,a)\in D} X(s,a)\Big)\Big]\\
&= \frac{1}{|D|^2}\E_{s,a}\Big[\sum_{(s,a)\in D} X(s,a)^2\Big] \\
& \quad\quad + \frac{2}{|D|^2}\E_{s,a,s',a'}\Big[\sum_{(s,a)\in D}\sum_{\substack{(s',a')\in\\  D-\{(s,a)\}}} X(s,a)X(s',a')\Big]\\
&= \frac{|D|}{|D|^2}\E_{s,a}\Big[X(s,a)^2\Big] \\
& \quad\quad + \frac{2}{|D|^2}\E_{s,a,s',a'}\Big[\sum_{(s,a)\in D}\sum_{\substack{(s',a')\in\\  D-\{(s,a)\}}} X(s,a)X(s',a')\Big] \\
&= \frac{1}{|D|}\E_{s,a}\Big[X(s,a)^2\Big] \\
& \quad\quad + \cancel{\frac{2}{|D|^2}\sum_{(s,a)\in D}\sum_{\substack{(s',a')\in\\  D-\{(s,a)\}}} \E_{s,a}\Big[X(s,a)\Big]\E_{s',a'}\Big[X(s',a')\Big]}
\end{split}
\quad
\begin{split}
\E_{s}[\text{MAC}^2] &= \E_{s}\Big[\Big(\frac{1}{|D|}\sum_{s\in D} Y(s)\Big)\Big(\frac{1}{|D|}\sum_{s\in D} Y(s)\Big)\Big] \\
&= \frac{1}{|D|^2}\E_{s}\Big[\sum_{s\in D} Y(s)^2\Big] \\
& \quad\quad + \frac{2}{|D|^2}\E_{s,s'}\Big[\sum_{s\in D} \sum_{\substack{s'\in \\D-\{s\}}} Y(s)Y(s')\Big]\\
&= \frac{|D|}{|D|^2}\E_{s}[Y(s)^2] \\
& \quad\quad + \frac{2}{|D|^2}\E_{s,s'}\Big[\sum_{s\in D} \sum_{\substack{s'\in \\D-\{s\}}} \E_{a}[X(s,a)]\E_{a'}[X(s',a')]\Big]\\
&= \frac{1}{|D|}\E_{s}[Y(s)^2] + \\
& \quad\quad + \cancel{\frac{2}{|D|^2}\sum_{s\in D} \sum_{\substack{s'\in \\D-\{s\}}} \E_{s,a}\Big[X(s,a)\Big]\E_{s',a'}\Big[X(s',a')\Big]}
\end{split}
\end{equation*}
\normalsize

By the assumption that $\widehat{Q}(s,a;\omega)$ is independent of $\widehat{Q}(s',a';\omega)$ for $(s,a) \neq (s',a')$, we can distribute the expectation through $\E[X(s,a)X(s',a')]$ in line 3 on the left, to obtain $\E[X(s,a)]\E[X(s',a')]$. In the last line, we can drop the second term in each expression, because they are the same. At this point we just need to compare $\E_{s,a}[X(s,a)^2)]$ vs. $\E_{s}[Y(s)^2]$. In order to make this comparison, we make use of Jensen's Inequality \cite{jensen1906fonctions}, which says that for a convex function $f$ and a vector $Z \in \mathbb{R}^n$:
$$ \E[f(Z)] \geq f(\E[Z]) $$

We note that $f(z)=z^2$ is convex, and as such, the following holds:
\begin{align*}
\forall_s \E_{a}[X(s,a)^2] \geq (\E_{a}[X(s,a)])^2 \implies \forall_s \E_{a}[X(s,a)^2] \geq Y(s)^2 \implies \E_{s,a}[X(s,a)^2] \geq \E_{s}[Y(s)^2]
\end{align*}

Thus, we can conclude that $\Var[\text{MAC}] \le \Var[\text{AC}]$. Moreover, since $f(z) = z^2$ is strictly convex, this inequality is strict as long as $a$ is not almost surely constant for a given state. That means for deterministic policies, we have $\Var[\text{MAC}] = \Var[\text{AC}]$, and for stochastic policies, $\Var[\text{MAC}] < \Var[\text{AC}]$.

}

\end{document}